\documentclass[letterpaper, 10pt, conference]{ieeeconf}      

\IEEEoverridecommandlockouts                              

\usepackage{graphicx}
\usepackage{epsfig}
\usepackage{amsmath}
\usepackage{amssymb}
\usepackage{amsfonts}
\usepackage{siunitx}
\usepackage{algorithm}
\usepackage{algpseudocode}
\usepackage{subcaption}
\usepackage{xcolor}

\usepackage{multicol, multirow}

\title{\LARGE \bf
A Heuristic Approach for Performance Tuning in RL-based Quadrotor Control via Reward Design and Termination Conditions 
}

\author{Fausto Mauricio Lagos Suarez$^{1}$, Akshit Saradagi, Vidya Sumathy, \\ and George Nikolakopoulos
\thanks{*This work has been funded by the European Union's Horizon Europe Research and Innovation Program, under the Grant Agreement No. 101119774 SPEAR.}
\thanks{*This research was conducted using the resources of High Performance Computing Center North (HPC2N). Additionally, the RL-training were enabled by resources provided by the National Academic Infrastructure for Supercomputing in Sweden (NAISS), partially funded by the Swedish Research Council through grant agreement no. 2022-06725.}
\thanks{$^{1}$Fausto Lagos is the corresponding author of the article
        {\tt\small faulag@ltu.se}. The authors are with the Robotics and AI group, in the Department of Computer Science, Electrical and Space Engineering at Luleå University of Technology, Sweden.}%
}

\begin{document}

\maketitle
\thispagestyle{empty}
\pagestyle{empty}

\begin{abstract}
Reinforcement learning (RL)-based quadrotor control policies have achieved impressive performance in tasks such as fast navigation in cluttered environments and drone racing, where the focus is on speed and agility. However, in several applications, such as infrastructure inspection, it is critical to achieve precise, controlled maneuvers with tunable performance. In this article, we present a novel heuristic approach to achieve tunable performance in RL-based Quadrotor control through reward design and termination conditions. We present a novel reward structure containing dual bandwidth exponentials that achieves a baseline critically damped response in setpoint tracking, with low steady-state errors. When trained with a Proximal Policy Optimization (PPO) algorithm, in conjunction with episode truncation conditions, the desired performance is achieved in 6 million time steps in a sample-efficient manner. In order to tune the performance about the baseline behavior, we present intuitive heuristic rules to adjust the reward weights and exponential coefficients to achieve faster (acrobatic-like) and slower (inspection-like) settling time performance, while retaining the baseline critically damped response and approximately 2\% steady-state error. We evaluate the three RL policies (baseline, acrobatic, and inspection) across 100 trials and show accurate and tunable performance in position and yaw tracking from random initial conditions, thereby demonstrating the effectiveness of the proposed heuristic approach.
\end{abstract}
\section{INTRODUCTION}
In the recent years, reinforcement Learning (RL) has been used to learn quadrotor navigation and control policies \cite{azar_drone_2021}, tackling challenging tasks such as flying through narrow, tilted, and moving gaps \cite{wang_narrow_gap}, vision-based safe navigation in cluttered environments \cite{kulkarni_reinforcement_2024} and surpassing human performance in drone racing \cite{kaufmann_champion-level_2023}. 
Despite such remarkable results, focused on pushing the limits of Quadrotor performance, in terms of speed and agility, providing guarantees for RL-control performance and safety, in the control theoretic sense, remains an open research direction. Moreover, typical RL training setups are rigid in terms of control over the resulting behaviors, and tuning the resulting behavior is extremely challenging. In applications such as industrial inspection, where it is critical to achieve precise, controlled maneuvers, it is essential to design RL-based controllers with tunable transient and steady-state performance.
%
%
\subsection{Related Works}
Among the wide range of quadrotor control algorithms that exist in literature \cite{khalid_control_survey}, the Proportional-Integral-Derivative (PID) control remains the most common approach \cite{lopez_pid_survey}. Due to nonlinear and strongly coupled dynamics, controlling all six Degrees Of Freedom (DOF) with a single PID structure is challenging. Consequently, cascade PID architectures that assign separate control structures to different DOF are widely adopted. One of the main reasons for the popularity of the cascade architecture is the ease in performance tuning and the existence of several intuitive tuning heuristics to influence control-theoretic metrics such as settling time, overshoot, and steady state error. In contrast, typical RL setups for training Quadrotor controllers involve complex reward functions which are difficult to iteratively tune, given the heavy time and compute costs involved in experience-based learning. 

In recent years, there have been efforts to design hybrid RL training setups that incorporate control-theoretic ideas and transient requirements. In \cite{Han_cascadeRL}, an RL-based cascade strategy uses six independent agents to control individual Quadrotor DOF and achieves step-response performance comparable to a classical PID controller. Along similar lines, \cite{mahran_SAC} combines an outer RL controller with an inner-loop PID that maps the RL-generated thrust vector to motor RPMs, indicating that a two-loop architecture can substantially improve performance in set-point tracking.
In \cite{kunapuli2025leveling}, Geometric Control (GC) is compared against an RL controller for agile trajectory tracking, showing that GC achieves slightly smaller steady-state error, whereas RL exhibits a better transient performance. Beyond the hybrid RL-control approaches, \cite{quan_RL_lipschitz} improves PPO training stability through Lipschitz-constrained policy regularization, yielding competitive transient metrics relative to PID and Model Predictive Control (MPC). Finally, \cite{Bernini_formal_metrics} analyses the design of a pure RL controller with formal performance metrics, emphasizing the role of hyperparameters, network architecture, and observation design in achieving step-response behavior comparable to classical controllers. 

\subsection{Contributions} It is worth noting that, even in control-RL hybrid approaches discussed so far, the focus is on training just one policy with some performance expectations and not on designing RL training setups that provide control over the performance of the policy. 

In this article, we present a novel heuristic approach to achieve tunable performance in RL-based Quadrotor control through reward design and termination conditions. A novel reward function that incorporates dual-bandwidth exponentials is proposed and achieves a critically damped baseline response in end-to-end Quadrotor setpoint tracking, with low steady-state errors, when trained with the PPO algorithm. In order to tune the performance about the baseline performance, we present intuitive heuristic rules to adjust the reward weights, exponential coefficients, and termination conditions to achieve faster (acrobatic-like) and slower (inspection-like) settling time performance, while retaining the baseline steady-state error of approximately 2\%. The effectiveness of the proposed heuristic in achieving tunable performance is statistically validated over 100 trials of the baseline, acrobatic, and inspection policies in the position stabilization task.
\section{PROBLEM FORMULATION} \label{sec:problem_formulation}
\subsubsection{Quadrotor system.} 
\begin{figure}[h!]
    \centering
    \includegraphics[scale=.35]{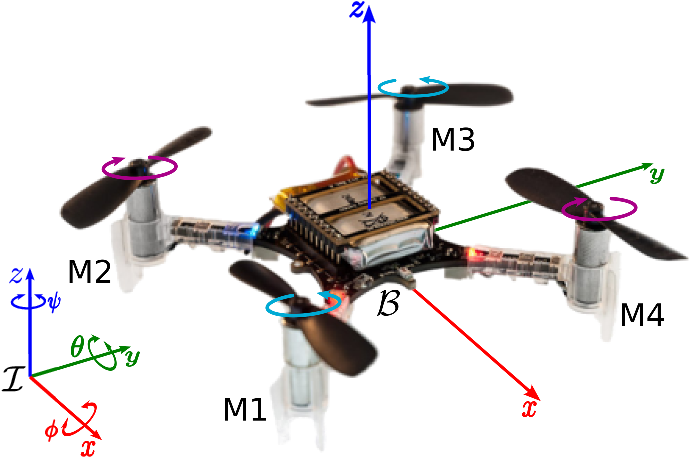}
    \caption{The Crazyflie 2.x Quadrotor, along with the body and inertial frames of reference and the rotational directions of its motors.}
    \label{fig:quadrotor_bodyframe}
\end{figure}

In this article, we consider the Crazyflie quadrotor \cite{Forster2015CrazyflieSysID} in $\times$ configuration shown in Fig. \ref{fig:quadrotor_bodyframe}. To have a singular free representation, we choose the unit Quaternions over Euler angles to represent the Quadrotor's 3D orientation. 

The quaternion rigid-body dynamics as in \cite{Greiff2017ModellingAC} with the $(x, y, z, w)$ quaternion convention is given by:
\begin{equation}
    \begin{cases}
        \overset{..}{p}_{I} = \frac{1}{m}\Im \left(q\otimes[\begin{array}{cc} T_{I} & 0\end{array}]^T\otimes q^\ast \right)-g\hat{z}_{I}-\frac{1}{m}D_{I}\dot{p}_{I} \\
        \dot{q} = \frac{1}{2}q\otimes [\begin{array}{cc}\omega_{B} & 0\end{array}]^T \\
        \dot{\omega}_B = I^{-1}_B(\tau_B-[\omega_B]_\times I_B\omega_B),
    \end{cases}
\end{equation}
where $p_I \in \mathbb{R}^3$ is the position of the quadrotor's Center Of Mass (COM) expressed in the inertial frame $I$, $q \equiv q_{BI} \in \mathbb{R}^4$ is the unit quaternion describing the orientation of the body frame $B$ with respect to $I$, stored as $q = [\begin{array}{cccc}q_x & q_y & q_z & q_w\end{array}]^T$, $\omega_B \in \mathbb{R}^3$ is the angular velocity of the body expressed in body frame $B$, $m \in \mathbb{R}_{>0}$ is the mass of the drone, 
$g$ is the gravitational acceleration magnitude, $\hat{z}_I \in \mathbb{R}^3$ is the unit vector of the global up/down axis used to apply gravity where up is $+z$, $I_B \in \mathbb{R}^{3\times 3}$ is the inertia matrix of the rigid body about its COM, expressed in $B$, $[\omega_B]_\times \in \mathbb{R}^{3\times 3}$ is the skew-symmetric matrix, $T_B \in \mathbb{R}^3$ is the total thrust force vector in $B$, $\tau_B \in \mathbb{R}^3$ is the control torque vector about body axes, and the term $-\frac{1}{m}D_I\dot{p}_I$ is a simple velocity-proportional gradient.
\subsubsection{Problem statement.} \label{sec:problem_statement}
Overall, the problem of designing a model-free RL training workflow to train an end-to-end Quadrotor stabilizing controller with tunable performance requirements is considered in this article. The classical step-response metrics are used to assess stabilization performance, as they provide a principled way to formalize desired closed-loop behavior and a standardized basis for controller evaluation. Specifically, motivated by the requirements in the infrastructure inspection scenario, this article considers a critically damped response in position stabilization, with low steady-state error as the baseline. This article seeks an RL training workflow that achieves the baseline, along with intuitive heuristics to tune the workflow parameters to enable faster (acrobatic) and slower (inspection) responses relative to the baseline, while retaining low steady-state errors. 
\section{RL setup for training Quadrotor stabilization policies} \label{sec:rl-framework}
\begin{figure}[h!]
    \centering
    \includegraphics[width=\linewidth]{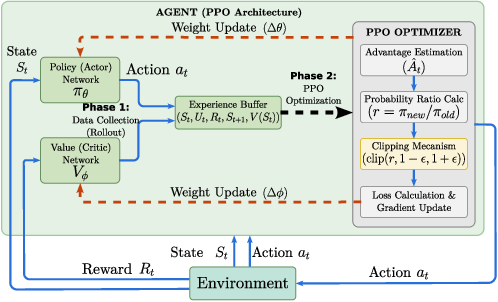}
    \caption{Reinforcement Learning setup and illustration of the functioning of the PPO algorithm in optimizing the policy during training. Solid blue lines represent the interaction loop, and dashed orange lines represent the internal PPO updates.}
    \label{fig:rl-framework}
\end{figure}

In this work, the RL environment is modeled as a Partially Observable Markov Decision Process (POMDP) and represented as a tuple of five elements $(\mathcal{S}, \mathcal{A}, \mathcal{P}, \mathcal{R}, \gamma)$, where $\mathcal{S}$ is the state (observation) space, $\mathcal{A}$ is the action space, $\mathcal{P}$ is the state transition probability, $\mathcal{R}$ is the reward function, and $\gamma \in (0, 1)$ is the discounting factor. Fig. \ref{fig:rl-framework} presents the overall RL learning loop, and the role of the Proximal Policy Optimization (PPO) algorithm in the policy learning process. The PPO learning algorithm is chosen in this work for its stability and sample-efficient design \cite{schulman_ppo}, which are desirable for our goal of developing an RL-based quadrotor controller with tunable performance conditions.
\subsection{Observation space}
The observation space of the agent is defined as a vector $\vec{\mathbf{o}}_t = [\begin{array}{ccccc}\tilde{\mathbf{e}}_{p, t} & \tilde{\mathbf{q}}_{e, t} & \tilde{\mathbf{v}_t} & \tilde{\omega}_t & \mathbf{a}_{t-1}\end{array}]^T \in \mathbb{R}^{17}$, where $\tilde{\mathbf{e}}_{p, t} \in \mathbb{R}^3$ is the position error with noise, $\tilde{\mathbf{q}}_{e, t} \in \mathbb{R}^4$ is the error quaternion with noise, $\tilde{\mathbf{v}_t} \in \mathbb{R}^3$ is the linear velocity with noise, $\tilde{\omega}_t \in \mathbb{R}^3$ is the angular velocity with noise, and $\mathbf{a}_{t-1} \in [-1, 1]^4$ is the last policy action.

We added Gaussian noise to all components of the observation space, except for the last four actions, to increase the robustness of the policy to real-time sensor noise, which is an essential consideration for real-time implementation. The magnitude of the added noise is as specified in Table \ref{tab:observation_noise}.
\renewcommand{\tabcolsep}{1mm}
\renewcommand{\arraystretch}{1.5}
\renewcommand{\arrayrulewidth}{1pt}
\begin{table}[h!]
    \centering
    \begin{tabular}{cc}
        \hline
        \textbf{Observation Component} & \textbf{Noise magnitude} \\
        \hline
        $\tilde{\mathbf{e}}_{p, t}$ & $\mathcal{N}(0,\sigma_p^2I_3)$, $\sigma_p = 10^{-3}$ \\
        $\tilde{\mathbf{q}}_{e, t}$ & $\mathcal{N}(0, \sigma_q^2I_4), \sigma_q = \num{2e-3}$ \\
        $\tilde{\mathbf{v}_t}$ & $\mathcal{N}(0, \sigma_v^2I_3), \sigma_v = 10^{-3}$ \\
        $\tilde{\omega}_t$ & $\mathcal{N}(0, \sigma_\omega^2I_3), \sigma_\omega = \num{2e-3}$ \\
        \hline
    \end{tabular}
    \caption{Magnitude of the Gaussian noise injected in each state component of the observation space.}
    \label{tab:observation_noise}
\end{table} 
%
%
\subsection{Action space}
This article aims to develop an end-to-end policy that directly maps the drone states to the raw control inputs. The action space is a vector $\vec{a} \in [-1, 1]^4$ of the normalized motor RPM values mapped in the simulation to the Crazyflie motors' RPM $\in (0, \num{24e3})$ using the linear approximation $RPM_i = \left.RPM_{Hover} \cdot \left(1 + \frac{a_i}{2}\right)\right|_{i=1}^4$
where $RPM_{Hover} = \sqrt{\frac{gm}{4k_f}}$, with $g$ the gravitational acceleration magnitude, $m$ the drone mass, and $k_f$ the torque constant. We use motor RPMs as the action space because of their versatility and independence of additional control structures.
\subsection{Reward design and termination conditions}
\subsubsection{Reward function} \label{sec:reward_function}
In an RL architecture, the reward term is one of the most important elements as it indirectly guides the policy to learn the desired behavior. The goal of RL algorithms is to train an agent to obtain the most cumulative reward over time. In this paper, a multi-component reward function is designed to incorporate a wide range of aspects crucial in learning stabilizing control policies for a Quadrotor. The proposed control-informed reward function is defined as follows:
\begin{multline}
    R(t) = s^R + \lambda_1\mathbf{e}_{xy}^R + \lambda_2\mathbf{e}_z^R + \lambda_3\mathbf{v}^R + \lambda_4\theta^R - \lambda_5\mathbf{a}_\Delta^P,
    \label{eq:reward_function}
\end{multline}
where the individual components are:   
\textbf{a. Survival reward:} We use $s^R$ as the survival reward \cite{Eschmann2021} to incentivize the agent to maintain its life, i.e., don't crash or terminate the learning episode.
\textbf{b. Position error reward ($\mathbf{e}_{xy}^R + \mathbf{e}_z^R$):} Introduces the classical cascade approach to handle altitude and horizontal position separately, the position error reward is defined as an exponential reward in the $(x, y)$ error plus an exponential reward in the $z$ error. 
\textbf{c. Linear velocity reward ($\mathbf{v}^R$):} Rewards the agent progressively while the linear velocity magnitude tends to zero. 
\textbf{d. Geodesic angle reward ($\theta^R$):}
We use the geodesic angle to represent the orientation error from the quaternion error. The geodesic angle represents the shortest angular distance between two 3D orientations (the target orientation and the drone's current orientation). We use the axis-angle representation of the error quaternion to calculate the geodesic angle as $\theta = 2\arctan2(||v||, w)$, where $v$ is the vector part, and $w$ the scalar part of the error quaternion. 
\textbf{e. Smoothness penalty ($\mathbf{a}_\Delta^P$):} It is defined as the weighted squared Euclidean norm of the difference between consecutive actions serving as an approximation of the control-action derivative, thereby introducing a damping term in the reward function and encouraging the neural network to learn derivative-like control behavior.
\subsubsection{Termination and truncation conditions}
The termination conditions are success or failure signals, whereas the truncation condition is the time horizon for the agent to survive. The stability of the learning process depends as much on the reward function as on the terminal and truncation conditions \cite{pardo_time_limits}, implying that unrealistic termination conditions reduce the learnability of the reward function. In this paper, we defined failure termination conditions as when the drone enters unsafe or undesired operating states, such as descending to less than 10 cm above the floor, or accelerating excessively, which may increase the likelihood of overshoot, particularly in obstacle-rich environments. 
\section{Heuristics for performance tuning} \label{sec:heuristic}
Mapping classic control specifications (such as steady-state error, settling time, and overshoot) to specific elements in the RL setup is particularly challenging. 
This is because, fundamentally, the RL objective of maximizing cumulative reward and the requirement of meeting time-domain performance specifications have to be reconciled. To overcome this challenge, this paper proposes a heuristic approach that formulates a control-informed reward function constrained by specific termination conditions and an intuition-guided tuning process (validated in section \ref{sec:heuristic}).
\subsection{Reward shaping}
The survival reward ($s^R$) tuning aims to balance action-taking with survival. Larger survival rewards ($s^R \approx 0.1$) are needed to encourage an agent to explore risky actions to respond faster. Conversely, when actions are constrained to be smooth and safer, $s^R$ can be reduced, preventing the agent from learning behaviors that lead to non-optimal rewards. The remaining components of the proposed reward shaping, except the smoothness penalty ($\mathbf{a}_\Delta^P$), are designed as exponential rewards with single or dual-bandwidth to address the trade-off between response time (settling time) and steady-state accuracy. A single bandwidth exponential reward is defined as $R = e^{-\delta x^2}$. In a dual-bandwidth reward $R = \alpha e^{-\delta_\alpha x^2} + \beta e^{-\delta_\beta x^2}$, the desired response is split into coefficients influencing early response ($\alpha$) and steady-state accuracy ($\beta$). In both cases, $\delta$ controls how soon the agent receives positive rewards, encouraging the agent to respond faster $\delta > 1$ or slower $\delta < 1$. Finally, the smoothness penalty ($\mathbf{a}_\Delta^P$) is tuned via its weight ($\lambda_5$), together with the angular velocity ($\omega$) termination conditions. Larger $\lambda_5$ along with larger $\omega$ constraints will over-penalize changes in consecutive actions, leading to large steady-state errors. 
%
\subsection{Termination conditions tuning}
This work considers five tunable termination conditions to constrain the quadrotor behavior and influence the control response. The $z$ termination condition is associated with the quadrotor altitude, informing the agent when it has crashed, influencing its survivability. The positional error ($p_e$), as well as the geodesic angle ($\theta_G$) termination conditions, prevent the agent from exploring actions that guide the agent to move away from the target. $\theta_G$ can also be split using Euler angles to control specifically roll and pitch rotations. The linear and angular velocities are bounded within specific ranges, which govern the magnitude of the action taken and influence the quadrotor's agility and critically damped response.
\section{Validation setup} \label{sec:validation}
To demonstrate the effectiveness of the proposed heuristics for tuning the rewards and the truncation conditions, in training quadrotor controllers with tunable performance, three policies were trained: a baseline policy, an acrobatic policy with faster response, and an inspection-like policy with slower response. In all cases, a steady-state error below $2\%$ of the target is desired.
Equations \eqref{eq:acrobatic_reward} to \eqref{eq:inspection_reward} present the reward shaping for the three trained policies, and Table \ref{tab:termination_conditions} shows the corresponding tuning of the termination conditions, highlighting the tuned parameters: red indicates the parameter increases w.r.t. the baseline \eqref{eq:baseline_reward} and blue the parameter decreases. 

\subsubsection{Acrobatic Policy} For the acrobatic policy \eqref{eq:acrobatic_reward}, the early response bandwidth of the $xy$ position error reward is increased, encouraging the agent to reduce the $xy$ error faster, while the importance of the linear velocity reward and the linear velocity termination condition are both relaxed, allowing the agent to move faster for a longer time. 

\subsubsection{Inspection Policy} In the inspection policy \eqref{eq:inspection_reward}, because the linear and angular velocity termination conditions are shrunk, the agent is guided toward smooth state changes. The survival reward is relaxed, thereby reducing the policy's likelihood of becoming trapped in a local minimum. The sharpness of the $xy$ position error reward is increased, and a dual-bandwidth $z$ position error reward is introduced to encourage accuracy, compensating for the required action softness and reduction in the linear and angular velocities boundaries. The importance and sharpness of the linear velocity are increased, inducing the agent to reduce the linear velocity to zero early during one episode, as well as the geodesic error reward. In the last component of the reward function, the action softness is relaxed, compensating for the reduction in the angular velocity boundary, along with the separated bounded roll and pitch termination conditions instead of the coupled geodesic angle termination condition.
\begin{multline} \label{eq:acrobatic_reward}
    R(t) = 0.05 + 0.25(0.6e^{-\textcolor{magenta}{4.0}||xy - xy^\ast||^2} + \\ 0.4e^{-20||xy - xy^\ast||^2}) + 0.25e^{-4||z - z^\ast||^2} + \\ \textcolor{cyan}{0.08}e^{-\textcolor{cyan}{0.5}||v||^2} + 0.15e^{-0.5\theta^2} - 0.1||a_t - a_{t - 1}||_2^2
\end{multline}
\begin{multline} \label{eq:baseline_reward}
    R(t) = 0.05 + 0.25(0.6e^{-0.5||xy - xy^\ast||^2} + \\ 0.4e^{-20||xy - xy^\ast||^2}) + 0.25e^{-4.0||z - z^\ast||^2} + \\ 0.1e^{-||v||^2} + 0.15e^{-0.5\theta^2} - 0.1||a_t - a_{t - 1}||_2^2
\end{multline}
\begin{multline} \label{eq:inspection_reward}
    R(t) = \textcolor{cyan}{0.01} + 0.25(0.6e^{-\textcolor{magenta}{4.0}||xy - xy^\ast||^2} + \\ 0.4e^{-\textcolor{magenta}{150}||xy - xy^\ast||^2}) + 0.25(\textcolor{teal}{0.6e^{-4.0||z - z^\ast||^2} +} \\ \textcolor{teal}{0.4e^{-150||z - z^\ast||^2}}) + \textcolor{magenta}{0.15}e^{-\textcolor{magenta}{1.5}||v||^2} + \\ \textcolor{magenta}{0.2}(\textcolor{teal}{0.6e^{-0.5\theta^2} + 0.4e^{-150\theta^2}}) - \textcolor{cyan}{0.02}||a_t - a_{t - 1}||_2^2
\end{multline}
\begin{table}[h!]
    \centering
        \begin{tabular}{cccccccc}
        \hline
        \multicolumn{1}{c}{\multirow{2}{*}{\textbf{Policy}}} & \multicolumn{7}{c}{\textbf{Termination conditions}} \\
        \multicolumn{1}{c}{}  & \multicolumn{1}{c}{$z$} & $p_e$ & $\theta_G$ & $\phi$ & $\theta$ & $v$ & $\omega$\\
        \hline
        Acrobatic & $\downarrow0.1$ & $\uparrow3.0$ & $\uparrow\pi$ & - & - & \textcolor{cyan}{$\uparrow1.0$} & $\uparrow530$ \\
        Baseline & $\downarrow0.1$ & $\uparrow3.0$ & $\uparrow\pi$ & - & - & $\uparrow 0.8$ & $\uparrow 530$ \\
        Inspection & $\downarrow0.1$ & $\uparrow3.0$ & - & \textcolor{magenta}{$\uparrow15^\circ$} & \textcolor{magenta}{$\uparrow15^\circ$} & \textcolor{magenta}{$\uparrow0.2$} & \textcolor{magenta}{$\uparrow115$} \\
        \hline
        \end{tabular}
    \caption{Termination conditions for both policies. The conditions are the minimum altitude $z$, the error position $e_p$, geodesic angle $\theta_G$, roll angle $\phi$, pitch angle $\theta$, linear velocity magnitude $v$, and angular velocity magnitude $w$. Blue indicates that the boundary was relaxed, while red indicates that the boundary was shrunk with respect to the baseline.}
\label{tab:termination_conditions}
\end{table}
\section{RL Training and Validation Results} \label{sec:results}
\subsubsection{Neural network architecture}
The actor (policy) neural network is a Multi-layered Perceptron (MLP) network with four layers: the first is a layer of 17 inputs (the observation space), the two hidden layers have 64 fully connected nodes with $\tanh$ activation functions, and the last layer is a four-node layer (the action space). The critic (value function) network has the same architecture, except for the last layer, which is a one-node layer yielding the output of the value function. 
%
\subsubsection{PPO hyperparameters}
For training all three policies, we use the same set of PPO hyperparameters as in Table \ref{tab:PPO_hyperparameters}.
\begin{table}[h!]
    \centering
    \begin{tabular}{ccccc}
        \hline 
        \textbf{Batch size} & \textbf{Learning rate} & \textbf{Steps} & \textbf{Epochs} & \textbf{Clip range} \\
        \hline
        128 & \num{2e-4} & 4096 & 12 & 0.15 \\
        \hline
    \end{tabular}
    \caption{The three policies were trained with the same set of PPO hyperparameters.}
    \label{tab:PPO_hyperparameters}
\end{table}
\subsubsection{Training}
For training, we used our own training framework built on top of Gym-PyBullet-Drones \cite{pybullet_drones}, a Gymnasium environment that uses PyBullet Physics as the physics engine and Stable-Baselines3 as the library of reliable RL algorithm implementations in PyTorch. Our training framework inherits the base training environments from Gym-PyBullet-Drones, adding new artifacts and utilities to introduce observation noise and to enhance the simulator to test the trained policy in scenarios with random initial states. The training was performed on a Compute Skylake Kebnekaise HPC2N node with one Intel Xeon Gold 6132 CPU and 192 GB of RAM, running 4 parallel environments.

\begin{figure}[h!]
    \centering
    \includegraphics[width=0.9\linewidth]{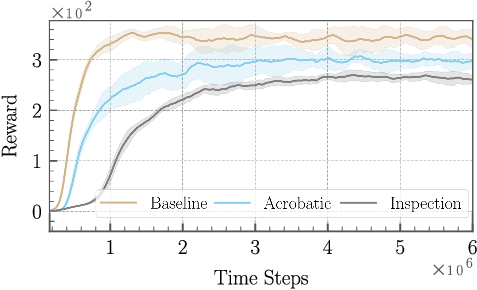}
    \caption{Rollout episode mean reward across 5 random seeds per trained policy.}
    \label{fig:rollout_reward}
\end{figure}

The physical parameters used to simulate the Crazyflie 2.1 in $\times$ configuration are explicitly shown in Table \ref{tab:crazyflie_parameters}. The control frequency corresponding to a complete interaction loop between the agent and the environment is 100 Hz.
\begin{table}[h!]
    \centering
    \begin{tabular}{ccc}
        \hline
        \textbf{Parameter} & \textbf{Notation} & \textbf{Value} \\
        \hline
        Mass &$m$ & 0.033 [$Kg$] \\
        Arm length &$d$ & \num{39.73d-3} [$m$] \\
        Torque constant & $k_f$ & \num{3.16d-10} \\
        Moment constant & $k_m$ & \num{7.49d-12} \\
        Propeller radius &$p$ & \num{23.1348d-3} [$m$] \\
         \hline
    \end{tabular}
    \caption{Physical parameters of the Crazyflie 2.1 in $\times$ configuration.}
    \label{tab:crazyflie_parameters}
\end{table} 
Each policy is trained for 6 million time steps. For each controller, we used five random seeds and report the training evolution in Fig. \ref{fig:rollout_reward} using the rollout episode mean reward. The small variance across seeds indicates that the proposed PPO setup is stable and not overly sensitive to reward tuning. As expected, the terminal conditions affect early learning dynamics: the inspection policy requires approximately the first 700.000 time steps to learn feasible survival behavior under stricter termination criteria before improving the return and converging to higher rewards.
\subsubsection{Evaluation.} 
To evaluate the proposed heuristic for learning quadrotor control under tunable performance requirements, we conducted 100 test trials, each lasting 10 s. The trials were initialized from random states: initial positions were sampled uniformly with $(x_0, y_0) \in [-2, 2]^2$ m, and $z_0 \in [0, 2]$ m, and initial attitude with $(\phi_0, \theta_0) \in [-15^\circ, 15^\circ]^2$, and $\psi_0 \in [-180^\circ, 180^\circ)$. Each trial targets a stable hover at $(x, y, z)^\ast = (0, 0, 1)$, with attitude $(\phi, \theta, \psi)^\ast = (0^\circ, 0^\circ, 0^\circ)$. We evaluate settling time, overshoot, and steady-state error for four variables: the three position components $(x, y, z)$, and the yaw angle $\psi$. We assess three performance behaviors: a baseline response, an acrobatic response with a shorter settling time than the baseline, and an inspection response with a slower, smoother transient response than the baseline. In all three policies, a steady-state error below 2\% of the target is expected. 

\begin{figure}[h!]
    \centering
    \begin{subfigure}[t]{\linewidth}
        \centering
        \includegraphics[width=0.9\textwidth]{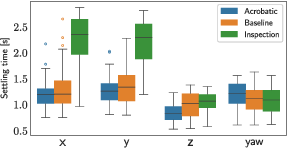}
        \label{fig:acrobatic_overshoot}
    \end{subfigure}
    \vfill
    \begin{subfigure}[t]{\linewidth}
        \centering
        \includegraphics[width=\textwidth]{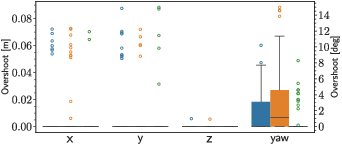}
        \label{fig:acrobatic_settling_time}
    \end{subfigure}
    \vfill
    \begin{subfigure}[t]{\linewidth}
        \centering
        \includegraphics[width=\textwidth]{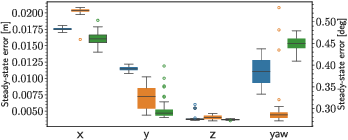}
        \label{fig:acrobatic_SSE}
    \end{subfigure}
    \caption{The three policies exhibit an critically damped response (center), the settling time (up), as expected, is consistently shorter in the acrobatic policy and larger in the inspection policy w.r.t the baseline. The three policies exhibit a steady-state error (down) within 2\% of the target.}
    \label{fig:boxplot}
\end{figure}

\begin{figure}[h!]
    \centering
    \includegraphics[width=\linewidth]{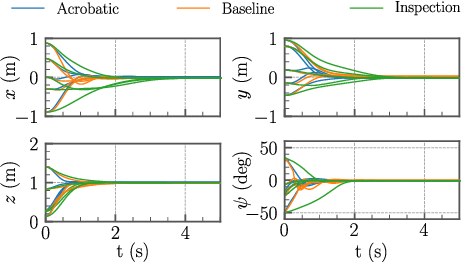}
    \caption{Performance profile comparison between the three trained policies in five random tests.}
    \label{fig:performance_profile}
\end{figure}

Figure \ref{fig:boxplot} reports these metrics for the three policies. The overshoot shown in Figure \ref{fig:boxplot} (center) for $x, y$, and $z$ exhibits zero Interquartile Range (IQR), indicating that at least 75\% of the test trials achieve the intended critically damped response. In contrast, yaw overshoot is larger, and, along with the outliers observed in the position variables, this can be attributed to the randomized initial conditions. Figure \ref{fig:boxplot} (up) shows that all trials satisfy the expected settling time for the three policies, demonstrating the convenience of the proposed heuristic in training RL-based controllers with different performance behaviors. Figure \ref{fig:boxplot} further shows accurate convergence to the target: steady-state errors remain within the prescribed 2\% band.
Figure \ref{fig:performance_profile} illustrates the performance profile of the three policies in five random tests. The three policies are remarkably accurate around the target, highlighting the reliability of the proposed heuristic in training RL-based controllers with different performance requirements.

\begin{figure}[h!]
    \centering
    \includegraphics[width=0.95\linewidth]{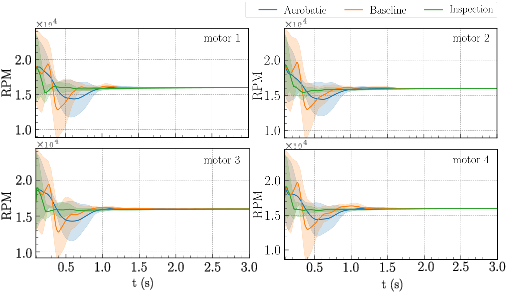}
    \caption{Motor RPMs average across 5 tests with random initialization.}
    \label{fig:rpms}
\end{figure}

Figure \ref{fig:rpms} presents the average motor RPMs for the three policies in 5 random tests. It is noticeable that the inspection policy requires less motor actuation and settles faster than the acrobatic and baseline policies, as expected by design.
%
\vspace{-8pt}
\section{Conclusions and Future Work} \label{sec:conclusions}
In this article, we presented a novel heuristic approach to achieve tunable performance in RL-based Quadrotor control through reward design and termination conditions. The central contribution is a intuitive heuristics for reward-shaping and termination-condition tuning which enables the training of controllers with different transient and steady-state characteristics. The learned policies perform position and attitude tracking by mapping observations consisting of position error, quaternion error, linear velocity, angular velocity, and the previous four actions directly to motor RPM commands. The proposed approach was validated by training three controllers, namely \emph{Acrobatic}, \emph{Baseline}, and \emph{Inspection}, corresponding to critically damped responses with different settling times, and a steady-state error requirement of 2\%. Using the same PPO hyperparameter configuration, all three policies showed similar learning efficiency and reached reward saturation at $6 \times 10^6$ time steps. Across 100 evaluation trials, the trained controllers consistently satisfied the target step-response metrics in terms of settling time, overshoot, and steady-state error. Robustness was further demonstrated under randomized initial conditions, as well as under Gaussian observation noise. Overall, the results show that the proposed heuristic constitutes a reliable and practical methodology for the design of RL-based quadrotor controllers with tunable stabilizing performance.

Future work will address sim-to-real transfer and experimental validation, with particular emphasis on domain randomization, latency, and onboard computational limitations.






\bibliographystyle{ieeetr}
\bibliography{2026/MED_references}

\end{document}